# Hierarchical Reinforcement Learning Framework for Stochastic Spaceflight Campaign Design[1]

Yuji Takubo[2], Hao Chen[2], and Koki Ho[3]

*Georgia Institute of Technology, Atlanta, GA, 30332*

**This paper develops a hierarchical reinforcement learning architecture for multi-mission spaceflight campaign design under uncertainty, including vehicle design, infrastructure deployment planning, and space transportation scheduling. This problem involves a high-dimensional design space and is challenging especially with uncertainty present. To tackle this challenge, the developed framework has a hierarchical structure with reinforcement learning (RL) and network-based mixed-integer linear programming (MILP), where the former optimizes campaign-level decisions (e.g., design of the vehicle used throughout the campaign, destination demand assigned to each mission in the campaign), whereas the latter optimizes the detailed mission-level decisions (e.g., when to launch what from where to where). The framework is applied to a set of human lunar exploration campaign scenarios with uncertain in-situ resource utilization (ISRU) performance as a case study. The main value of this work is its integration of the rapidly growing RL research and the existing MILP-based space logistics methods through a hierarchical framework to handle the otherwise intractable complexity of space mission design under uncertainty. We expect this unique framework to be a critical steppingstone for the emerging research direction of artificial intelligence for space mission design.**

## Nomenclature

| | | |
|---|---|---|
| $\mathscr{A}$ | = | set of arcs |
| $\boldsymbol{a}_S$ | = | vehicle design actions, kg |
| $\boldsymbol{a}_I$ | = | space infrastructure deployment action, kg |
| $\boldsymbol{c}$ | = | commodity cost coefficient matrix |
| $c'$ | = | vehicle cost coefficient |
| $\mathcal{C}_c$ | = | continuous commodity set |
| $\mathcal{C}_d$ | = | discrete commodity set |
| $\boldsymbol{d}$ | = | mission demand, kg |
| $H$ | = | concurrency constraint matrix |
| $\mathcal{J}$ | = | mission planning objective (mission cost), Mt |
| $\mathcal{M}$ | = | memory buffer |
| $NN_{vehilce}$ | = | set of nodes |
| $\mathcal{N}$ | = | set of nodes |
| $\boldsymbol{q}$ | = | stochastic mission parameter vector |
| $r$ | = | reward |
| $\boldsymbol{s}$ | = | state |
| $\mathcal{T}$ | = | set of time steps |
| $\mathcal{U}$ | = | set of state variables |
| $\mathcal{V}$ | = | set of vehicles |
| $V$ | = | value function of vehicle design (estimated campaign cost from the second to the final mission) |
| $\hat{v}$ | = | estimated total campaign cost, Mt |
| $W$ | = | set of time windows |
| $\boldsymbol{x}$ | = | commodity outflow variable |
| $\beta$ | = | basis function |
| $\theta$ | = | basis function coefficient |
| $\lambda$ | = | propellant mass fraction |
| $\Gamma$ | = | number of space missions |
| $\pi$ | = | policy |

*Subscripts*

| | | |
|---|---|---|
| $i$ | = | node index |
| $j$ | = | node index |
| $p$ | = | commodity index |
| $t$ | = | time step index |
| $u$ | = | state variable index |
| $v$ | = | vehicle index |

$\tau$     =   space mission index

# I. Introduction

**A**S an increasing number of space exploration missions are being planned by NASA, industry, and international partners, managing the complexity and uncertainty has become one of the largest issues for the design of cislunar and interplanetary missions. Particularly, in a multi-mission space campaign, each mission is highly dependent on one another, which can cause new challenges that would not be seen for conventional mission-level design. First, the interdependency between the missions can lead to the cascading of the technical or programmatic uncertainties of one mission to other missions in the campaign, similar to the "cascading failure" [1] or the bullwhip effect [2] in supply chain problems. To counter the undetermined factors, it is necessary to consider stochasticity in large-scale space campaigns for safe human space exploration. Additionally, as the technologies for in-situ resource utilization (ISRU) or on-orbit services mature, the demands of future space missions are fulfilled not only from the earth but also from the pre-positioned facilities in space [3,4]; this adds complexity to the problem as both deployment and utilization need to be considered for these infrastructure elements for a campaign-level analysis. Finally, assuming a family of common vehicle (spacecraft) design is used for the campaign, we need to consider the trade-off of infrastructure deployment and vehicle design used for the campaign, as the larger vehicle can deploy more ISRU plants but requires a higher cost. The vehicle design is also dependent on the basic mission demand such as a habitat or other fundamental facilities, and so we need an integrated framework that considers the entire resource supply chain. Previous studies have not succeeded in formulating an efficient optimization architecture that can address all these challenges at the same time.

In response to these challenges, we develop a new optimization framework based on hierarchical reinforcement learning (HRL). The idea behind the proposed hierarchical structure is to use reinforcement learning (RL) to optimize campaign-level decisions and use network-based mixed-integer linear programming (MILP) to optimize the detailed mission-level decisions. The campaign-level decisions include the design of the vehicle used throughout the campaign (i.e., spacecraft design) and the determination of the destination demand assigned to each mission in the campaign (i.e., space infrastructure deployment strategy), each of which can be trained with separate levels of RL agents. The mission-level decisions can be made for each mission, including when to launch what from where to where (i.e., space transportation scheduling), which can be optimized using a MILP-based dynamic generalized multi-commodity flow formulation. All these levels of decisions are interdependent on each other, and the proposed RL-MILP hierarchical structure of the decisions enables this integrated optimization under uncertainty to be solved effectively. As a case study, the framework is applied to a set of human lunar exploration campaign scenarios with uncertain in-situ resource utilization performance.

The value of this paper is in its novel framework to solve campaign-level space mission design problems. As reviewed in the next section, although numerous optimization-based approaches have been proposed to solve this problem, all of them have challenges in their scalability for realistic problems under uncertainty. The proposed framework introduces a completely new way to tackle this challenge, leveraging the rapidly advancing RL and MILP in a unique way. The proposed framework is generally compatible with any RL algorithms. In the later case study, a comparison of different state-of-the-art RL algorithms for the proposed RL-MILP framework is conducted and their performances are analyzed. With a growing number of high-performance RL methods being developed every day, the framework is expected to be even more powerful. We believe that the proposed method for the large-scale spaceflight campaign can open up a new future research direction of artificial intelligence for space mission design.

The remainder of this paper proceeds as follows. Section II mentions the literature review for the space logistics optimization frameworks and RL. Section III introduces the proposed methodology in detail. Section IV describes the problem setting for the case studies and analyzes the results. Finally, Section V concludes the analysis and refers to potential future works.

## II. Literature Review

### A. Space Logistics Optimization

The state-of-the-art space logistics analysis methods are based on time-expanded network modeling. Multiple studies have treated campaign-level mission planning such as SpaceNet [5], Interplanetary Logistics Model [6], and a series of network-based space logistics optimization frameworks based on the generalized multicommodity network flow and MILP [7–13]. The MILP-based optimization formulation theoretically guarantees the global optima for any deterministic problem scenarios. However, as the complexity of the campaign scenario increases, the computation time increases exponentially. More critically, this formulation cannot handle the uncertainties; naively introducing the uncertainties using stochastic programming can quickly increase the numbers of variables and constraints, making the problem intractable. Several papers attempted to consider the uncertainties in the space mission planning optimization [14–17]; however, they are designed for specific mission architectures or with known decision rules; none of them can be applied to a general spaceflight campaign design.

### B. Reinforcement Learning

Reinforcement learning (RL) is an algorithm of machine learning. In general, an RL agent sophisticates a policy $\pi: \mathcal{S} \to A$ that determines an action $\boldsymbol{a} \in A$ which maximizes the reward $r$ under a given state $\boldsymbol{s} \in \mathcal{S}$. Since the agent can autonomously learn from its trials, it has broad applications from robotics [18], board games [19], or feedback control [20]. This method is also called Approximate Dynamic Programming (ADP) in the field of mathematical optimization [21]. There have also been studies on the optimization of large-resource allocation [22] or the determination of locomotive design or scheduling in a multicommodity flow network [23,24], although none of them can handle the complexity for optimizing the infrastructure deployment and the vehicle design concurrently under general uncertainty.

There are various algorithms proposed to solve RL problems. Most model-free RL algorithms can be categorized as an on-policy algorithm and an off-policy algorithm [25]. On-policy algorithms (e.g. State-Action-Reward-State-Action (SARSA)[26], Trust Region Policy Optimization (TRPO)[27], Proximal Policy Optimization (PPO)[28]) train the agent from the latest policy which is used for the action selection. For each episode, experiences $(\boldsymbol{s}, \boldsymbol{a}, r, \boldsymbol{s}')$ are created as training data based on the latest policy, and the policy is updated based on these experiences. On the other hand, off-policy algorithms (e.g. Q-learning[29], Deep Deterministic Policy Gradient (DDPG)[30], Soft Actor-Critic (SAC)[31]) train the agent based on the data in the replay buffer. The buffer contains not only the experiences based on the latest learned policy but also those based on the past policy, and the agent extracts the training data from the buffer. Off-policy algorithms are efficient in terms of data-sampling as they can reuse past experiences, which generally has competence in complicated real-world problems, whereas on-policy algorithms have to create data sets for each episode. However, since the data extracted from the buffer can contain experiences based on the different policies, off-policy algorithms can potentially deteriorate the learning process, creating a high sensitivity to the hyperparameters.

Also, there are two policy types that can be used for the RL agent: deterministic policy [32] and stochastic policy [33]. A deterministic policy (e.g. Q-learning, DDPG, Twin-Delayed DDPG (TD3)[34]) returns the same action when given the state. In contrast, a stochastic policy (e.g. PPO, SAC) returns the same probability distribution of the mapping of state to the action; the agent can return different actions when given the same state under the stochastic policy. The stochastic policy is expected to perform well under the uncertain process while it can complicate the training process, and so the applicability of the method highly depends on the characteristics of the problem.

The proposed framework is compatible with all of these algorithms, particularly in the state-of-the-art actor-critic RL framework. In the later case study, we choose the representative algorithms based on these categorizations.

To deal with a problem that requires high complexity, hierarchical architectures for RL, or HRL [35] have been proposed. HRL decouples complicated actions into sets of actions, thus making it easier for the agent to learn the optimal policies. One of the most fundamental architectures of HRL is the Options Framework [36,37], in which a higher level of abstract actions are regarded as options (sub-goal), and a detailed action is chosen using an intra-option policy to achieve the option. Another fundamental architecture of HRL is the MAXQ framework [38]. It decomposes tasks into high-level and low-level action spaces. The Q-function of the low-level action space is defined as a sum of the value of the action in the low-level task (sub-tasks) and the supplemental value of the low-level action for the high-level task (parent-task). By inserting a lower Markov Decision Process (MDP) into a high-level MDP, the MAXQ framework successfully evaluates the decoupled actions in the sub-task. However, these existing methods do not apply to our space mission design problem because: (1) we do not have a clear policy model that can be used to connect the high-level and low-level tasks, and (2) the reward of the low-level task cannot be decoupled from that of the high-level task.

Inspired by the idea of the HRL and leveraging the unique structure of the space mission design problem, this paper develops a new framework that uses the idea of HRL in combination with network-based MILP modeling to handle the complexity in the stochastic spaceflight campaign design problem.

### III. Methodology

We consider a large-scale space campaign that comprises multiple missions (i.e., launches of multiple vehicles in multiple time windows), where we need to satisfy certain payload delivery demands to the destinations at a (known) regular frequency (e.g., consumables and equipment to support a habitat). If each mission has the same payload demand, a trivial baseline solution would be to repeat the same missions independently of each other every time the demand emerges. However, this is not necessarily the optimal solution because we also have the technology for infrastructure (e.g., ISRU) which requires a large cost for initial deployment but can be used to reduce costs of later missions. Whether such infrastructure can reduce the total campaign cost or not needs to be analyzed at the campaign level. Furthermore, we assume the vehicle design (i.e., sizing) needs to be fixed before the campaign, and that design is used for all vehicles used in the campaign. These assumptions are made for simplicity and can be relaxed when needed for various applications. The main objective of the stochastic spaceflight campaign optimization is to find the set of vehicle design and infrastructure deployment plan that minimizes the (expected) total campaign cost, as well as the detailed logistics of the commodity flow of the mission, under uncertainties (e.g., the uncertain performance of the ISRU infrastructure). In this paper, the objective is to minimize the sum of the initial mass at low-earth orbit (IMLEO) at each mission; other cost metrics can also be used if needed.

This section describes the developed methodology in detail. We first introduce a bi-level RL, which considers the RL and network-based MILP, and then extend to a more advanced tri-level RL, which adds another RL agent for vehicle design as another level. Then, we will explain each level of the framework in more detail.

## A. Architectures for HRL

### 1) Bi-Level Reinforcement Learning Architecture

The challenge of using RL for spaceflight campaign design is its large action space; the actions for space mission design contain every detailed logistics decision, including when to launch what from where to where over a long time horizon, which can make the learning process computationally intractable.

One solution to this challenge is to use the network-based MILP formulation to determine the detailed mission-level decisions, while the RL agent is used to provide high-level guidance. This architecture is referred to as a bi-level RL architecture.

Each of the levels in the bi-level RL architecture is organized as follows:

First, the RL agent determines the campaign-level infrastructure deployment action plan at each mission (i.e., ISRU deployment plan) as well as the vehicle design (i.e., spacecraft). Here, the high-level structure of the problem is modeled as a Markov Decision Process (MDP), where each mission is regarded as one step in the decision-making process. In this architecture, the actions are defined as the infrastructure deployment plan for each mission and the vehicle design used for the campaign, and the states are defined as the available resources at the key nodes (e.g., lunar surface) after each mission. The rewards can be defined by the reduction of IMLEO compared with the baseline.

Second, given the infrastructure deployment action plan and vehicle design from the RL agent, the space transportation scheduling is determined by the network-based space logistics optimization method, which is formulated as MILP. The calculated mission cost is fed back to the RL agent as a reward. Note that each execution of this MILP only needs to optimize one mission logistics given the infrastructure deployment action plan as the demand, which can be completed with a small computational cost.

By iterating the action determined by the RL agent and MILP-based space transportation scheduling, the RL agent learns the optimal vehicle design and the infrastructure deployment plan. Fig. 1 represents the overview of the bi-level RL framework. $\boldsymbol{a}_{I,\tau}$ denotes the infrastructure deployment action at mission $\tau$, and $\boldsymbol{a}_S$ denotes a vehicle design action; $r$ denotes a reward, which is the mission cost; $\boldsymbol{s}$ indicates a state vector.

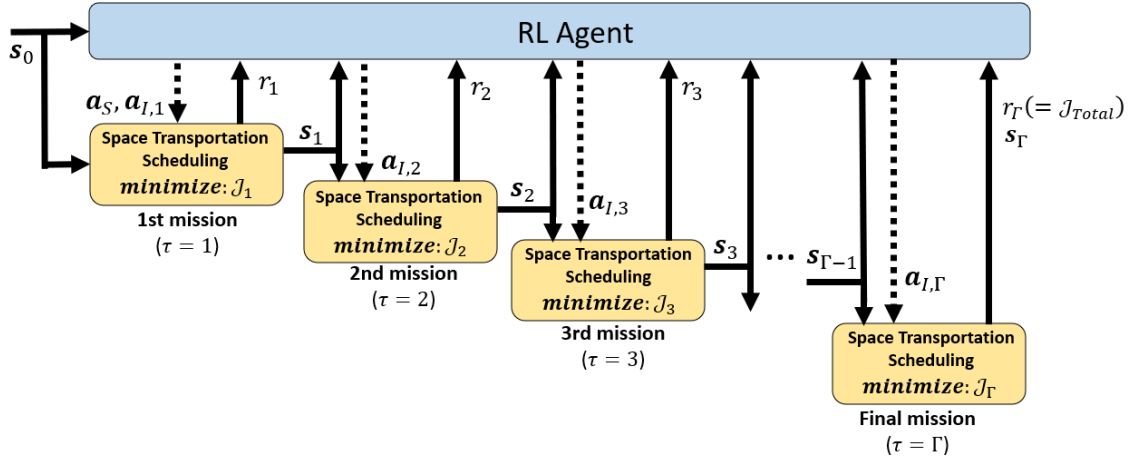

**Fig. 1 Bi-level RL architecture for space campaign design**

The advantage of the bi-level RL architecture is its simplicity and generality. By splitting the decision-making agent into the RL agent and space transportation scheduling, the RL agent can theoretically include any high-level decisions (what commodities we need to bring, or how much commodities we need to bring to achieve the mission) so that space transportation scheduling can automatically optimize the low-level decisions (how to transport the commodities).

### 2) Tri-Level Reinforcement Learning Architecture

While bi-level RL architecture decouples the entire problem into high-level and low-level decision-making problems, we can further leverage our knowledge of the problem structure to refine the architecture and facilitate the learning process. Specifically, we can separate vehicle design decisions from infrastructure deployment decisions because the former is considered to be constant throughout the campaign whereas the latter is varied at every mission in the campaign. Thus, we propose a tri-level RL architecture, where we separate the vehicle design as another level on top of the infrastructure deployment and space transportation scheduling. Each of the levels in the tri-level RL architecture is organized as follows:

The first level in the tri-level RL is the vehicle design agent, which determines the vehicle design. In our formulation, this vehicle design is optimized together with space transportation scheduling of the first mission to ensure that we find a feasible vehicle design at least for the first mission. If the demand of each later mission is the same or less than the that of the first mission (which is true in later case studies), the found vehicle design is feasible for the entire campaign. When determining the vehicle design, not only the influence on the first mission but also that on the future mission should be considered. Therefore, we add the value function approximation (VFA) term, which takes vehicle design parameters as arguments, only to the objective function at the first mission and expect the VFA expresses the value of the vehicle design in the future. If the VFA accurately expresses the cost of the second through the final mission, we can obtain the optimal vehicle design even at the beginning of the campaign.

Secondly, the infrastructure deployment agent receives the information of vehicle design and status quo of infrastructure deployment (i.e., state) and returns the infrastructure deployment plan at each mission (i.e., action). Note that, unlike the bi-level RL architecture, the action for this MDP does not include the vehicle design, because the vehicle design is considered at the above level.

Finally, the space transportation scheduling optimizes the mission-level logistics and calculates the cost of the mission given the infrastructure deployment action plan from the RL, which is fed back to the two RL agents discussed above. In the same way as the bi-level RL architecture, this optimization is formulated as a MILP.

By iterating these episodes, we can sophisticate the spaceflight campaign design. The HRL solves the circular reference of the design variables, especially the interconnection of vehicle design and infrastructure deployment, by separating the design domains into two RL agents and one MILP optimization method. The abstract hierarchical architecture is shown in Fig. 2. The infrastructure deployment agent iteratively outputs the action for each mission in an episode (campaign), and the vehicle design agent iteratively outputs vehicle design parameters at the beginning of each campaign.

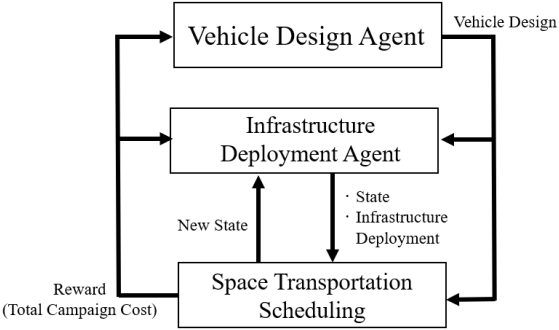

**Fig. 2 Abstract Hierarchy of the space campaign design architecture**

**B. Algorithms for the HRL**

This subsection introduces the detailed concepts and algorithms for the HRL. The explanation in this subsection is based on the tri-level RL architecture because it is a more advanced version, although a similar set of algorithms can also be used for bi-level RL architecture as well; the only difference is that there would be no vehicle design agent, and instead, vehicle design actions would be provided by the infrastructure deployment agent. In the following, we introduce each level of the proposed HRL-based architecture: vehicle design agent, infrastructure deployment agent, and space transportation scheduling.

### 1) Vehicle Design Agent

The vehicle design is determined by a Value-based RL algorithm at the campaign level. Even though the vehicle design has to be determined at the beginning of the campaign, it should be chosen with consideration of the future influence in the campaign. To account for the influence of the future mission, we set the value function $V(\boldsymbol{a}_S)$ to represent the mission cost from the second to the final mission; $\boldsymbol{a}_S$ indicates vehicle design action, and this is regarded as state variables for the vehicle design agent. If we can completely predict the future cost of the campaign based on the vehicle design, we can choose the vehicle design which minimizes the total campaign cost even at the beginning of the campaign. The general formulation of the VFA using a neural network can be shown as follows:

$$V(\boldsymbol{a}_S) = NN_{vehicle}(\boldsymbol{a}_S)$$

By updating the neural network of the vehicle design agent until its convergence, we can determine the optimal vehicle sizing. Because the vehicle design must be optimized together with the space transportation scheduling of the first mission to guarantee feasibility, the objective at the first mission of the campaign can be written as

$$\hat{v}_{Total}(\boldsymbol{a}_S) = \mathcal{J}_1(\boldsymbol{a}_S) + NN_{vehicle}(\boldsymbol{a}_S)$$

where $\mathcal{J}_1$ is the cost of the first mission, and $\hat{v}_{Total}$ is the estimated total campaign cost with the VFA term. Note that (1) when the vehicle design is determined (i.e., optimized) through the space transportation scheduling, the detailed mission operation of the first mission is simultaneously optimized, and (2) infrastructure deployment must be chosen before the vehicle design is optimized via space transportation scheduling, so infrastructure deployment agent will choose the infrastructure deployment at the first mission before getting knowledge of the vehicle design.

A pseudo-code of the vehicle design agent is shown below. We denote the actual total campaign cost by $\mathcal{J}_{Total}$.

---

Vehicle Design Agent Pseudo Code

---

Initialize value function approximation neural network $NN_{vehicle}^0$.

Initialize the iteration counter $m = 1$ and set the maximum episodes $M$.

Set the baseline demand and supply for each mission.

**for** $m = 1: M$ **do**

Solve the approximation problem for the first mission (with vehicle design) by choosing the optimal $\boldsymbol{a}_S^m$ which minimizes $\hat{v}_{Total}^m$, where $\hat{v}$ is the approximated total campaign cost.

$$\hat{v}_{Total}^m(\boldsymbol{a}_S^m) = \mathcal{J}_1^m(\boldsymbol{a}_S^m) + NN_{vehicle}^{m-1}(\boldsymbol{a}_S^m)$$

Update state variable $\boldsymbol{a}_S^m$.

Obtain the actual total campaign cost at the end of the campaign, $\mathcal{J}_{Total}^m(\boldsymbol{a}_S^m)$.

Update $NN_{vehicle}^{m-1}$ to $NN_{vehicle}^m$ by gradient descent on

$$\left(\hat{v}_{Total}^m(\boldsymbol{a}_S^m) - \mathcal{J}_{Total}^m(\boldsymbol{a}_S^m)\right)^2$$

**end for**

---

To make $NN_{vehicle}$ optimizable for the space transportation scheduling that is solved via MILP, we define the value function approximation based on the linear combination of the basis functions as follows.

$$V(\boldsymbol{a}_S) = NN_{vehicle}(\boldsymbol{a}_S) = \sum_{u \in \mathcal{U}} \theta_u \beta_u(\boldsymbol{a}_S)$$

where $u$ is the index of state variables; $\beta_u(\boldsymbol{a}_S)$ is the basis function that extracts specific pieces of information from each state $\boldsymbol{a}_S$; $\theta_u$ is the corresponding coefficient. For the basis function, normalization of vehicle design parameters is performed.

If there are $u$ types of defining parameters of the vehicle design, we can write $\boldsymbol{a}_S$ as $\boldsymbol{a}_S = [a_1, a_2, \dots, a_u]^T$ and corresponding coefficient vector $\boldsymbol{\theta} = [\theta_1, \theta_2, \dots, \theta_u]^T$. At iteration $m$, we get a set of state variables $\mathcal{A}_S^m = [\boldsymbol{a}_S^1, \boldsymbol{a}_S^2 \dots, \boldsymbol{a}_S^m]^T$ and corresponding observed actual mission cost from the second to the final mission $\mathcal{J}_{2:\Gamma}^m = [\mathcal{J}_{2:\Gamma}^1, \mathcal{J}_{2:\Gamma}^2, \dots, \mathcal{J}_{2:\Gamma}^m]^T$, where $\Gamma$ is the total number of missions for one campaign. The linear approximation of the value function can be expressed as

$$V^m(\boldsymbol{a}_S^m) = NN_{vehicle}^{m-1}(\boldsymbol{a}_S^m) = \boldsymbol{\theta}^{m^T} \boldsymbol{a}_S^m$$

By using the least square method, $\boldsymbol{\theta}^m$ can be expressed as

$$\boldsymbol{\theta}^m = (\mathcal{A}_S^{m^T} \mathcal{A}_S^m)^{-1} \mathcal{A}_S^{m^T} \mathcal{J}_{2:\Gamma}^m$$

However, as the scale of the problem gets larger, it will be expensive to calculate $(\mathcal{A}_S^{m^T} \mathcal{A}_S^m)^{-1}$. Thus, we instead use the iterative update of $\boldsymbol{\theta}^m$ through the recursive least square method [21]. Here, if we define $\mathcal{B}^m = (\mathcal{A}_S^{m^T} \mathcal{A}_S^m)^{-1}$ as the matrix inverse at iteration $m$ and approximated $\mathcal{B}^m$ and $\boldsymbol{\theta}^m$ can be found as following recursions.

$$\mathcal{B}^m = (I - \frac{\mathcal{B}^{m-1} \boldsymbol{a}_S^m \boldsymbol{a}_S^{m^T}}{1 + \boldsymbol{a}_S^{m^T} \mathcal{B}^{m-1} \boldsymbol{a}_S^m}) \mathcal{B}^{m-1}$$

$$\boldsymbol{\theta}^m = \boldsymbol{\theta}^{m-1} - \mathcal{B}^m \boldsymbol{a}_S^m (\boldsymbol{\theta}^{m-1^T} \boldsymbol{a}_S^m - \mathcal{J}_{2:\Gamma}^m)$$

Note that in general, gradient descent can be used to update the neural network of the vehicle design agent. However, we use the least square method in this case since $NN_{vehicle}$ is a linear combination of the state variables.

To sum up, the vehicle design agent has a form of Value-based RL, which decides the vehicle design by combining the neural network and space transportation scheduling of the first mission. The neural network of the vehicle design agent expresses the value function as a function of the vehicle design, while the space transportation scheduling chooses the optimal vehicle design as well as other detailed mission-level scheduling decisions. After the vehicle design and the space transportation scheduling of the first mission are determined, the vehicle design is evaluated through the subsequent missions. At the end of the episode, the vehicle design agent receives the total campaign cost as a reward and updates its neural network.

### 2) Infrastructure Deployment Agent

The deployment of infrastructure for resource utilization is optimized through an RL algorithm. At mission $\tau$, the agent determines the amount of infrastructure deployment as an action $\boldsymbol{a}_{I,\tau}$ based on the state $\boldsymbol{s}_\tau$. Note that the vehicle design is not considered as an action for the infrastructure deployment agent in the tri-level RL architecture, and is instead regarded as states. Also, any other state variables may be added to expand the scope of the mission interdependencies we want to investigate. After the chosen action $\boldsymbol{a}_{I,\tau}$ is executed, the agent obtains a scalar reward $r_\tau$ and the new state $\boldsymbol{s}_\tau$ under a probability $P(\boldsymbol{s}_\tau|\boldsymbol{s}_{\tau-1}, \boldsymbol{a}_{I,\tau})$. The reward and the new state are returned by the space transportation scheduling (see the next subsection). The infrastructure deployment agent improves its policy to maximize the sum of the reward through the campaign.

As explained in the previous subsection, the vehicle design is determined after the infrastructure deployment at the first mission is determined. From the second mission, the infrastructure deployment agent regards the vehicle design as a part of the state and returns the infrastructure deployment for each mission. Therefore, a zero vector is assigned to the vehicle design at the first mission as a state.

The values of the stochastic mission parameters $\boldsymbol{q}$ (e.g., infrastructure performance) are chosen based on probability distributions at the beginning of each episode. Initially, $\boldsymbol{q}$ is set as a zero vector because we do not know the exact values of these parameters at the beginning. As we start to observe the infrastructure operations from the second mission, we gain information about $\boldsymbol{q}$ as part of the state. Since the infrastructure deployment agent iteratively trains its policy, it can accept different values of states for each episode, which is how the uncertainty is considered in this optimization method. Also, for the algorithms which use mini-batch learning, this method enables the agent to stabilize the learning process and to be durable to the outliers which are optimized with the extreme values of the stochastic parameters.

In this paper, a reward at a certain mission is defined based on the difference between the baseline mission cost (i.e., the cost of a single mission without infrastructure deployment) and the mission cost with the infrastructure deployment, which is calculated by the space transportation scheduling. Note that in the vehicle design agent, space transportation scheduling is used as both a decision-making agent and environment, and it is used only as an environment in the infrastructure deployment agent. In a scenario that comprises $\Gamma$ missions, the reward at mission $\tau$ is calculated as follows.

$$r_\tau = \begin{cases} \displaystyle\sum_{k=1}^{\tau-1} r_k - 1 = \sum_{k=1}^{\tau-1} \frac{\mathcal{J}_{base} - \mathcal{J}_k}{\mathcal{J}_{base}} - 1 & \text{if infeasible} \\[3ex] \displaystyle\sum_{k=1}^{\tau} r_k \;=\; \sum_{k=1}^{\tau} \frac{\mathcal{J}_{base} - \mathcal{J}_k}{\mathcal{J}_{base}} & \text{else if } \tau = \Gamma \\[3ex] \qquad\quad 0 & \text{else} \end{cases}$$

where $\mathcal{J}_{base}$ is the baseline mission cost calculated by MILP, and $\mathcal{J}_\tau$ is the cost of the mission $\tau$ based on the decisions performed by the agents. If the optimized mission cost is lower than the baseline mission cost, the reward will gain a positive reward and vice versa. Note that zero rewards are returned to the infrastructure deployment agent except for the last mission because the objective of this optimization is the minimization of the total mission cost, and the rewards at the middle point of the campaign have no meaning compared to the overall cost savings of the campaign. Furthermore, depending on the infrastructure deployment strategy, some vehicle designs can make the space transportation scheduling problem (introduced in the next subsection) infeasible because they cannot satisfy the mission demand; this can happen during the training process if the given infrastructure deployment plan is too aggressive. (Note that even when the original problem is feasible, infeasibility can be encountered during the training depending on the chosen infrastructure deployment plan.) If an infeasible infrastructure deployment is returned, a large negative reward is returned to the agent, and the episode will be terminated so that a new campaign design will be attempted. However, if the campaign is terminated before the final mission, the agent cannot return the cost from the second to the final mission $\mathcal{J}_{2:\Gamma}$, which is required to update the vehicle design agent. Therefore, if the campaign is terminated at mission $\tau$, the cost from the mission $\tau$ to the final mission $\mathcal{J}_{\tau:\Gamma}$ is substituted with the baseline cost $(\Gamma - \tau + 1) * \mathcal{J}_{base}$, and the total cost is then fed to the vehicle design agent.

The developed general framework can be integrated with any RL algorithms $\mathbb{Q}$: on-policy and off-policy. The comparison between these methods is evaluated later with the case study. The generalized pseudo-code for the infrastructure deployment agent is shown below.

---

**Infrastructure Deployment Agent Pseudo Code**

---

Initialize the RL algorithm $\mathbb{Q}$.

**for** each iteration **do**
    Initialize the state $\boldsymbol{s}_0$.
    Choose the stochastic mission parameters $\boldsymbol{q}$ from the probability distributions.
    **for** $\tau = 1$: $\Gamma$ **do**
        $\boldsymbol{a}_{I,\tau} \sim \pi(\boldsymbol{a}_{I,\tau} | \boldsymbol{s}_{\tau-1})$
        $\boldsymbol{s}_\tau \sim P(\boldsymbol{s}_\tau | \boldsymbol{s}_{\tau-1}, \boldsymbol{a}_{I,\tau})$
        Obtain the reward $r_\tau$.
        $\mathcal{M} \leftarrow \mathcal{M} \cup \{(\boldsymbol{s}_{\tau-1}, \boldsymbol{a}_{I,\tau}, r_\tau, \boldsymbol{s}_\tau)\}$
    **end for**

**for** each training step **do**

    Extract the data for learning.

    Perform one step of training based on the algorithm $\mathbb{Q}$.

  **end for**

**end for**

---

### 3)    Space Transportation Scheduling

In this subsection, we introduce the network-based space logistics optimization based on the MILP formulation, which serves as the lowest level of an optimization method in the HRL architecture. Given the vehicle design and the infrastructure deployment for every single mission, this method solves the space mission planning problem to satisfy the demands of each mission, such as infrastructure deployment requests or crews. This formulation considers the problem as a time-expanded generalized multicommodity network flow problem [7,8] based on graph theory, where planets or orbits are represented by nodes, and trajectories of transportation are represented by arcs. In this formulation, all crew, vehicles, propellant, and other payloads are regarded as commodities flowing along arcs.

For the formulation of this mission planning framework, the decisions to be made during the space missions are defined as follows.

$\boldsymbol{x}_{vijt}$ = Commodity outflow variable: the amount of the outflow of each commodity from node $i$ to $j$ at time $t$ by vehicle $v$. Each component is a nonnegative variable and can be either continuous or integer (i.e., discrete) depending on the commodity type; the former commodity set (i.e., continuous commodity set) is defined as $\mathcal{C}_c$, and the latter commodity set (i.e., discrete commodity set) is defined as $\mathcal{C}_d$. If there are $p$ types of commodities, then it is a $p \times 1$ vector. Vehicle is also regarded as a part of commodity.

Also, we define the parameters and sets as follows.

$\mathcal{A}(\mathcal{V}, \mathcal{N}, \mathcal{N}, \mathcal{T})$ = Set of arcs.

$\mathcal{N}$ = Set of nodes. (index: $i, j$)

$\mathcal{T}$ = Set of time steps. (index: $t$)

$\mathcal{V}$ = Set of vehicle. (index: $v$)

$\boldsymbol{d}_{it}$ = Demand or supply of missions at node $i$ at time $t$. Demand is negative and supply is positive. ($p \times 1$ vector)

$\boldsymbol{c}_{vijt}$ = Commodity cost coefficient matrix.

$\mathcal{C}_c$ = Continuous commodity set.

$\mathcal{C}_d$ = Discrete commodity set.

$\Delta t_{ij}$ = Time of flight (TOF) along arc $i$ to $j$.

$Q_{vij}$ = Commodity transformation matrix.

$H_{vij}$ = Concurrency constraint matrix.

$W_{ij}$ = Time window(s) of a mission from node $i$ to $j$.

Along with the defined notations above, the mission planning architecture after the first mission can be written as the following optimization problem.

Minimize:

$$\mathcal{J} = \sum_{(v,i,j,t) \in \mathcal{A}} \boldsymbol{c}_{vijt}^T \boldsymbol{x}_{vijt} \tag{1}$$

Subject to:

$$\sum_{(v,j):(v,i,j,t) \in \mathcal{A}} \boldsymbol{x}_{vijt} - \sum_{(v,j):(v,i,j,t) \in \mathcal{A}} Q_{vji} \boldsymbol{x}_{vji(t-\Delta t_{ji})} \leq \boldsymbol{d}_{it} + \boldsymbol{a}_{I,it} \quad \forall t \in \mathcal{T} \quad \forall i \in \mathcal{N} \tag{2a}$$

$$H_{vij} \boldsymbol{x}_{vijt} \leq \boldsymbol{0}_{l \times 1} \quad \forall (v,i,j,t) \in \mathcal{A} \tag{2b}$$

$$\begin{cases} \boldsymbol{x}_{vijt} \geq \boldsymbol{0}_{p \times 1} & \text{if } t \in W_{ij} \\ \boldsymbol{x}_{vijt} = \boldsymbol{0}_{p \times 1} & \text{otherwise} \end{cases} \quad \forall (v,i,j,t) \in \mathcal{A} \tag{2c}$$

Where:

$$\boldsymbol{x}_{vijt} = \begin{bmatrix} x_1 \\ x_2 \\ \vdots \\ x_p \end{bmatrix}_{vijt}, \quad \begin{array}{l} x_n \in \mathbb{R}_{\geq 0} \;\; \forall n \in \mathcal{C}_c \\ x_n \in \mathbb{Z}_{\geq 0} \;\; \forall n \in \mathcal{C}_d \end{array} \quad \forall (v,i,j,t) \in \mathcal{A}$$

Equation (1) represents the objective function. It returns the total campaign cost as a sum of each commodity flow solution. Both the vehicle design agent and infrastructure deployment agent use this function to update their networks.

Equation (2a) is the mass balance constraint, which ensures that the commodity outflow is always smaller or equal to the sum of commodity inflows minus mission demands. $\boldsymbol{d}_{it}$ represents the baseline mission demands (or supplies) that only depend on mission scenarios in node $i$ at time $t$; $\boldsymbol{a}_{I,it}$ is the demand vector of infrastructure deployment, determined by the infrastructure deployment agent. Again, demand is negative in $\boldsymbol{d}_{it}$ and $\boldsymbol{a}_{I,it}$. Additionally, $Q_{vij} \boldsymbol{x}_{vijt}$ identifies the commodity inflow from node $i$ to node $j$ after the commodities flow along the arc.

Equation (2b) represents the concurrency constraint which limits the upper bound of the commodity flow based on the design parameters of the vehicle. In the equation, we assume that there are $l$ types of constraints.

In this paper, we set the upper bound of the commodity flow limited by the payload and propellant capacities as the only concurrency constraints.

Equation (2c) represents the time window constraint. As both the interval of the launch from the earth and the time length (days) spent on the transportation of each arc are specified as the mission is planned, the commodity flow has to be operated only during the time window assigned to each mission.

For the first mission, the objective function will be the sum of $\mathcal{J}_1$ and the VFA term. Also, spacecraft design parameters $\boldsymbol{a}_s$ are regarded as optimized variables. In this paper, we use a nonlinear vehicle sizing model developed by Taylor et al., in which the structure mass (dry mass) can be expressed as a function of the payload capacity and the fuel capacity [6]. We apply the piecewise linear approximation to recast the nonlinear function as a binary MILP formulation. Details of the constraints are in Ref. [9].

After the capacities and structure mass of the vehicle are determined at the first mission, the vehicle design is fixed for the rest of the campaign, and the design parameters are passed from the space transportation scheduling section to the infrastructure deployment agent as state variables.

### 4) Space Campaign Design Framework

By incorporating all methods discussed above, the whole framework of the (tri-level) HRL-based campaign design architecture can be formulated. In this integration, we introduce a set of two hyperparameters $n_1, n_2$ to represent when the learning starts during the training process. This is because off-policy algorithms usually require a "warm-up" to fill the memory buffer with transition data. $n_1$ and $n_2$ are used to represent the number of the initial iterations used for this "warm-up" for infrastructure deployment agent and vehicle design agent, respectively. For on-policy algorithms, the learning process starts from the first episode, $n_1 = n_2 = 0$.

The detailed flowchart of the tri-level optimization architecture is shown in Fig. 3. The pseudo-code for the integrated framework, which follows the flowchart, is shown as follows.

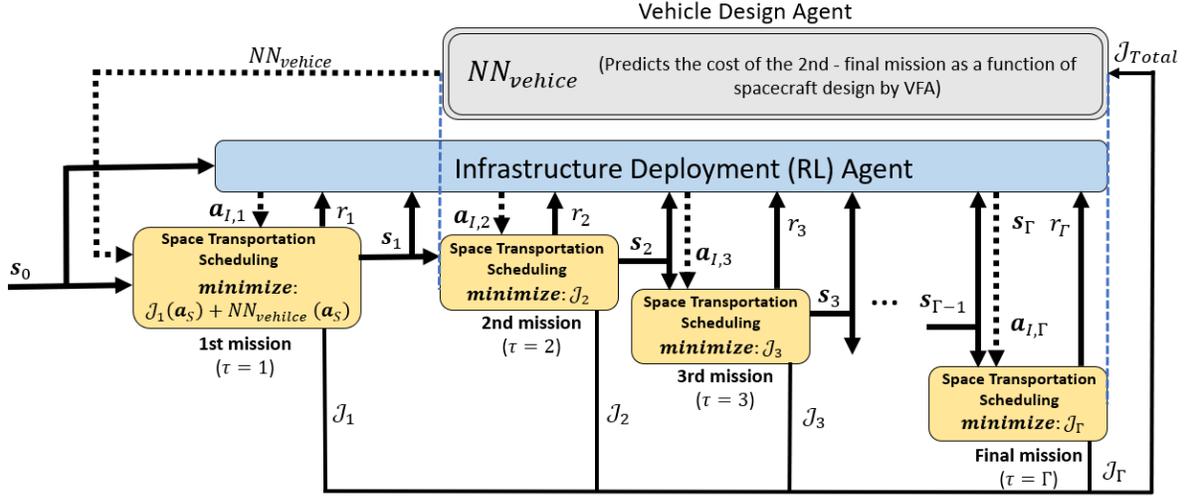

**Fig. 3 Tri-Level RL architecture overview for space campaign design**

---

**HRL-Based Campaign Design Framework Pseudo Code**

---

Given: A RL algorithm $\mathbb{Q}$ for Infrastructure agent          e.g. PPO, TD3, SAC

Infrastructure deployment agent: Initialize $\mathbb{Q}$ and the memory buffer $\mathcal{M}$.

Vehicle design agent: Initialize the neural network $NN_{vehicle}^0$ for VFA of the vehicle design.

Set the baseline demand and supply for each mission in the campaign, set the total number of missions in a campaign, $\Gamma$.

Initialize iteration counter $m = 1$, set the maximum episodes $M$.

Set the starting episode for the training of the agents, $n_1, n_2$.

**for** $m = 1$: $M$ **do**    ## loop for a total episode of training

    Initialize the state $s_0$.

    Choose the stochastic mission parameters $\boldsymbol{q}$ from the probability distributions.

    **for** $\tau = 1$: $\Gamma$ **do**    ## loop for a single campaign

        Choose $\boldsymbol{a}_{I,\tau}^m$ based on $\mathbb{Q}$.

        **if** $\tau = 1$ **then**

            Obtain the vehicle design $\boldsymbol{a}_S^m$ by solving the integrated problem with both transportation scheduling and vehicle designing.

$$\hat{v}_{Total}^m(\boldsymbol{a}_S^m) = \mathcal{J}_1^m(\boldsymbol{a}_S^m) + NN_{vehicle}^{m-1}(\boldsymbol{a}_S^m)$$

        **else**

            Use $\boldsymbol{a}_S^m$ and solve the transportation scheduling for one mission without vehicle designing.

        Obtain the reward $r_\tau$ and observe a new state $s_\tau$.

        Store the transition.

$$\mathcal{M} \leftarrow \mathcal{M} \cup \{(s_{\tau-1}, \boldsymbol{a}_{I,\tau}^m, r_\tau, s_\tau)\}$$

        **end if**

        **if** $m > n_1$ **then**

            Sample a minibatch from $\mathcal{M}$ and perform an update of $\mathbb{Q}$.

        **end if**

    **end for**

    Update vehicle design $\boldsymbol{a}_S^m$.

Obtain the total mission cost of the campaign. $\mathcal{J}_{Total}^m(\boldsymbol{a}_S^m)$.

**if** $m > n_2$ **then**

    Update $NN_{vehicle}^{m-1}$ to $NN_{vehicle}^m$ by gradient descent.

**else**

    $NN_{vehicle}^m = NN_{vehicle}^{m-1}$ (no update)

**end if**

**end for**

---

## IV. Case Study: Human Lunar Exploration Campaign

To examine the performance of the proposed architecture for large-scale space campaign designs, a multi-mission human lunar exploration campaign is set up in this section. In this case study, the extraction of water from the moon is assumed as the ISRU mechanism, where the electrolyzed water is used for the propellant as hydrogen and oxygen. We compare the performances of the representative RL algorithms for the infrastructure deployment agent. In Section IV.A, we describe the scenarios of the space campaign and the selection of the representative RL algorithms. Section IV.B elaborates on the results and analysis of the optimization done by each method.

### A. Problem Setting

This campaign model is regarded as a network flow problem that consists of the Earth, low earth orbit (LEO), low lunar orbit (LLO), and the Moon as nodes. Fig. 4 shows $\Delta V$ and the transportation time of flight (TOF) of each arc.

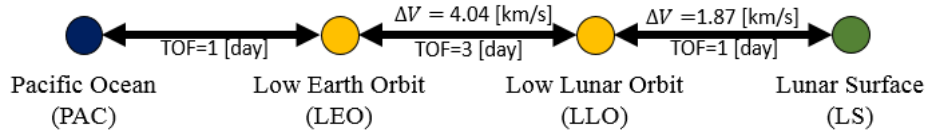

**Fig. 4 A multi-mission human lunar campaign**

The proposed campaign composes multiple year-long mission cycles, all of which have repeated baseline demands of crew, habitat/equipment, and returned samples. For each year, the launch window of the spacecraft from the earth to the moon opens at day 352; the crew stays on the moon for 3 days; the return window of the spacecraft from the moon to the earth opens at day 360; and the crew must be back on earth by day 365. These conditions correspond to equation (2c) that defines the launch window of the spacecraft. Also, the arcs where spacecraft are allowed to stay (i.e., set of arcs $\mathcal{A}$ in the space transportation scheduling) for the five-mission campaign are shown in Fig. 5. Similar rules apply to the other cases with different numbers of missions as well, and the time scale is not proportional to the actual timeline of the missions.

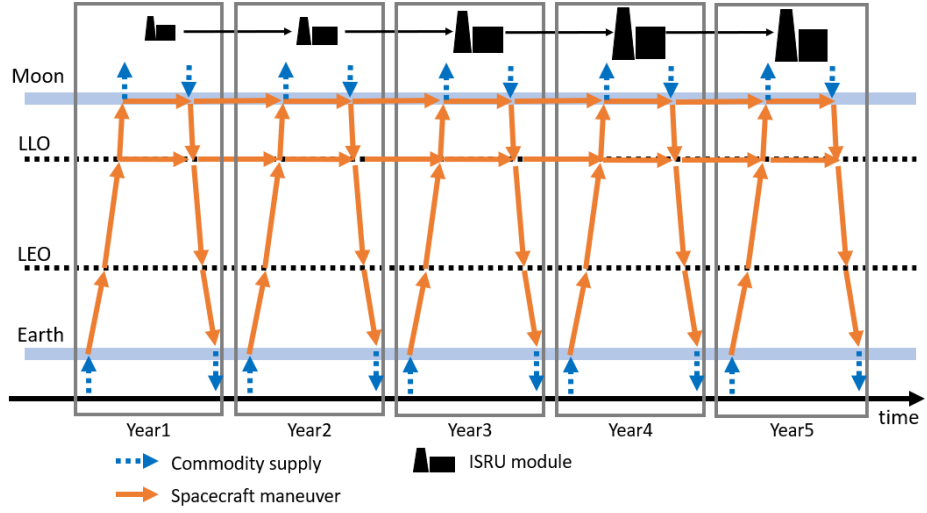

**Fig. 5 Possible arcs for the spacecraft for the five-mission campaign**

Table 1 represents the basic mission demands and supplies that each space mission has to satisfy with the corresponding time. The positive values in the supply column indicate the supply, and the negative values represent the demand at the node. The supply and the demand can be coupled: for example, sending the crew on day 352 from the earth can be regarded as a supply from the earth, which satisfies the demand (mission requirement) on the moon on day 357. Also, we assume that the earth can provide an infinite amount of commodity supplies at any time.

Furthermore, Table 2 shows the assumptions and parameters of the mission operation. As we introduce the water electrolysis ISRU model for the scenario, the propellant is also fixed as liquid oxygen (LOX) and liquid hydrogen (LH2), which has a specific impulse $I_{sp} = 420$ s and a structural fraction of the spacecraft propellant tank $\alpha = 0.079$ [39]. For each mission, 2,500kg of the lunar sample and other equipment are expected to be returned to the earth from the moon, and we set the upper bound of ISRU deployment in each mission as 5,000 kg. Additionally, we assume that both ISRU and spacecraft require a constant rate of maintenance. For ISRU, the maintenance facility, which is 5% of the total ISRU plant mass, is required for each year; for spacecraft, the maintenance materials, which are 1% of spacecraft structural mass, are expected for each flight (a maneuver from a node to another).

In this case study, the state variable $\boldsymbol{s}_\tau$ comprises the mission index $\tau$, the amount of deployed infrastructure at that time $\boldsymbol{s}_{I,\tau}$, the performance information about the infrastructure $\boldsymbol{q}$, and the vehicle design $\boldsymbol{a}_S$ (i.e., $\boldsymbol{s}_\tau = (\tau, \boldsymbol{s}_{I,\tau}, \boldsymbol{q}, \boldsymbol{a}_S)$), where $\boldsymbol{s}_0 = (0, \boldsymbol{0}, \boldsymbol{0}, \boldsymbol{0})$. Also, the action $\boldsymbol{a}_S$ comprises the payload capacity and propellant capacity of the spacecraft, while $\boldsymbol{a}_{I,\tau}$ is a scaler variable that indicates the mass of the deployed ISRU module at mission $\tau$.

**Table 1 Basic mission demands and supplies for each year**

| Phase | Payload Type | Node | Day | Supply |
|---|---|---|---|---|
| Go to Moon | Crew, # | Earth | 352 | + (Crew Number) |
| | | Moon | 357 | - (Crew Number) |
| | Habitat & equipment, kg | Moon | 357 | - (Habitat & Equipment) |
| | | Earth | All the time | $+\infty$ |
| | ISRU plants, and Propellant, kg | Earth | All the time | $+\infty$ |
| Back to Earth | Crew, # | Moon | 360 | + (Crew Number) |
| | | Earth | 365 | - (Crew Number) |
| | Return mass (Samples & materials), kg | Moon | 360 | +2500 |
| | | Earth | 365 | -2500 |

**Table 2 Assumptions and parameters for the mission operation**

| Parameter | Assumed value |
|---|---|
| Spacecraft propellant type | LH2/LOX |
| Spacecraft propellant $I_{sp}$, s | 420 |
| Types of Spacecraft designed, # | 1 |
| Number of vehicles for each type, # | 4 |
| Crew mass (including space suit), kg/person | 100 [9] |
| Crew consumption, kg/day/person | 8.655 [9] |
| Spacecraft maintenance rate, structure mass/flight | 1% [9] |
| ISRU maintenance rate, system mass/year | 5% |

Even though there has been significant progress in the research of ISRU in the last decade, there is still a large uncertainty in the performance of ISRU modules. This is because there are many technological means to extract oxygen such as extracting hydrated minerals from regolith, collecting water ice, or implementing ilmenite reduction [3,40]. Given the relatively low maturity level of these technologies and the highly dynamic and hostile operational environment, the exact ISRU productivity is often unknown beforehand. Therefore, large uncertainty exists in the ISRU productivity [41]. Additionally, during the operation, there is a considerable possibility that the productivity of the ISRU module will decay over time. If we think of a campaign with five missions, for example, the ISRU module deployed at the first mission has to be operated for four years on the extreme environment of the moon; there may be a failure of components that needs maintenance, or inevitable decay of productivity. The rate of decay itself is an uncertain parameter, which needs to be considered in the design.

To sum up, a number of uncertain factors can significantly affect the performance of ISRU, which can potentially lead to the failure of the space mission. In this case study, we define the production rate and decay rate of the ISRU module as normal distributions $N(\mu, \sigma^2)$. The lower bound of the ISRU productivity and the decay rate are truncated to zero by the definition of the parameter.

To examine the effectiveness and robustness of the proposed method over a variety of instances, we introduce ten scenarios of the campaign. The total number of missions, crew number, the supply of habitat and equipment, ISRU production rate, and ISRU decay rate are varied for each campaign scenario. The parameters for each scenario are shown in Table 3. Note that only ISRU parameters are regarded as stochastic parameters in this case study, but in general applications, any stochastic parameters can be integrated into the infrastructure deployment agent as states. The stochastic parameters are chosen at the beginning of each campaign (episode) and are constant until the end of it.

**Table 3 Campaign scenarios**

| Scenario | Number of Total Mission | Crew Number | Habitat & equipment [kg] | ISRU Production Rate (> 0) [kg-water/year /kg-plant mass] | ISRU Decay Rate (> 0) [%/year] |
|----------|-----|----|-------|-----------------|------------------|
| A | 2 | 6 | 5,000 | $N(5, 1.5^2)$ | $N(10, 10^2)$ |
| B | 3 | 6 | 5,000 | $N(5, 1.5^2)$ | $N(10, 10^2)$ |
| C | 4 | 6 | 5,000 | $N(5, 1.5^2)$ | $N(10, 10^2)$ |
| D | 5 | 6 | 5,000 | $N(5, 1.5^2)$ | $N(10, 10^2)$ |
| E | 4 | 6 | 5,000 | $N(10, 3^2)$ | $N(10, 10^2)$ |
| F | 5 | 6 | 5,000 | $N(10, 3^2)$ | $N(10, 10^2)$ |
| G | 4 | 6 | 5,000 | $N(5, 1.5^2)$ | $N(5, 5^2)$ |
| H | 5 | 6 | 5,000 | $N(5, 1.5^2)$ | $N(5, 5^2)$ |
| I | 5 | 12 | 10,000 | $N(5, 1.5^2)$ | $N(10, 10^2)$ |
| J | 5 | 12 | 10,000 | $N(10, 3^2)$ | $N(10, 10^2)$ |

Since random parameters are fed to the agent during the training of the RL agents, it is unfair to use the result of a final iteration of the training process when comparing the performances of the HRL-based method. To perform a fair comparison, we add a testing phase separately after the training phase with new stochastic parameter sets. In the testing phase, the total campaign cost is calculated under the same 128 stochastic cases (ISRU production rate and decay rate) through the Γ-mission MILP formulation and the average of them is regarded as the test result of the campaign cost. Note that the ISRU deployment in the first mission and the vehicle design are the same for all stochastic cases regardless of the stochastic parameters in the testing phase because the information about the uncertain parameters is only observable after the first mission.

As mentioned in Section III. B 2), any model-free RL algorithms can be applied to the infrastructure deployment agent. We introduced two categorizations (On-policy vs. Off-policy, and deterministic policy vs. stochastic policy) of RL algorithms, where the most model-free RL algorithms can be categorized into four groups. For the performance comparison, we chose state-of-the-art RL algorithms for each category, as shown in Table 4: PPO, TD3, and SAC. Note that an On-policy algorithm with a deterministic policy is a possible option; however, as discussed in Ref. [32], poor performance has been reported because the agent cannot learn from the data which contains a lot of the same experience sequences. Thus, we do not adopt the representative algorithm for this category of RL.

**Table 4 Representative RL algorithms and categorization**

|  | On-policy | Off-policy |
|---|---|---|
| Deterministic Policy | -- | TD3 |
| Stochastic Policy | PPO | SAC |

The hyperparameters of the RL algorithms are tuned independently of the testing dataset, and their values are listed in Appendix A. All numerical optimizations in this paper are performed by Python using Gurobi 9.0 solver on an i9-9940X CPU @3.3GHz CPU with RTX 2080 Ti and 64GB RAM. For the implementation, RL algorithms are based on Stable Baselines [42].

## B. Results and Discussion

Two comparison studies are set up to examine the effectiveness of the proposed optimization methods. First, we perform the architectural comparison between the bi-level and tri-level RL approaches. Then, by introducing the superior architecture, we compare the state-of-the-art MILP-based method [9] and the proposed HRL method with three representative RL algorithms.

When comparing the RL algorithms, the reproducibility of the results must be considered. It is well known that the same RL algorithm with the same hyperparameters behaves differently due to initial random seeds, and many algorithms are susceptible to hyperparameters. These factors make RL algorithms difficult to reproduce similar results [43]. To avoid the influence of stochastic effects, all trials of RL-based methods shown in this subsection are run multiple times under different initial random seeds. Both the best and average of the results are important: the best optimization results will be the most practical solution in the actual designing process under the given computation time, while the average and variance indicate the reproducibility of the results.

### 1) Bi-level RL vs. Tri-level RL

First, the results of the comparison between the two proposed architectures (bi-level and tri-level) are presented in Table 5, which represents the IMLEO for each campaign scenario. We run five trials for each optimization, and the best trial among them are presented in the table. For both architectures, TD3 is used for the infrastructure deployment agent as the RL algorithm. Additionally, the same number of the total timesteps are set for the two architectures for a fair comparison.

**Table 5 Architectural comparison (best-trial)**

| Scenario | Campaign cost (IMLEO), Mt | |
|---|---|---|
| | Bi-level | Tri-level |
| A | 1423.8 | 964.0 |
| B | 1790.6 | 1455.0 |
| C | 2534.7 | 1858.9 |
| D | 3338.7 | 2284.0 |
| E | 2689.3 | 1671.0 |
| F | 3282.9 | 2034.9 |
| G | 2564.5 | 1822.3 |
| H | 2732.9 | 2276.9 |
| I | 4262.1 | 3683.0 |
| J | 3857.9 | 3437.7 |

We can confirm that the tri-level RL architecture significantly outperforms the bi-level RL architecture, returning the campaign designs with smaller IMLEOs. One RL agent has to decide the optimal vehicle design in the bi-level RL architecture, and it is complicated because a single agent has to take the balance of infrastructure deployment and vehicle sizing at the same time. On the contrary, the tri-level RL architecture can optimize vehicle design after the infrastructure deployment is determined. Additionally, since the vehicle design can be optimized by space transportation scheduling by introducing VFA of the vehicle design, we can guarantee the feasibility of the found vehicle design (if there exists one) at the first mission and thus improve the learning efficiency; this enables more optimal vehicle design than the bi-level RL method under the same computation time.

For the following experiments, the tri-level RL architecture is adopted.

### 2) HRL vs. State-of-the-Art MILP

Next, we compare the performance of the HRL-based method with three RL algorithms for the infrastructure deployment agent against the state-of-the-art MILP-based method.

We first run the HRL-based methods five times and the best result among the five trials is compared. At the same time, the performances are compared with the deterministic MILP-based method. Since the MILP formulation cannot consider the randomness of the parameters, it has to adopt the worst scenarios for the stochastic parameters (zero ISRU productivity in this case). Any other scenario can potentially lead to an infeasible solution due to the overly optimistic assumptions; for example, a deterministically designed mission operation that assumes the best or mean ISRU productivity would be infeasible if the ISRU productivity is worse than that. Note that the worst zero-ISRU situation does not necessarily mean that each mission is completely independent; the MILP-based method still allows the reuse of the vehicles or deployment of propellant depots [44] if it finds these solutions preferred in terms of the cost metric. Also, for reference, the semi-worst case, which adopts the 5th percentile of the ISRU productivity and 95th percentile of the decay rate, is optimized by the MILP-based method; note that this case would not return feasible results for the worst scenario (i.e., no ISRU) and therefore is only included for reference and cannot be compared fairly with the RL results. The maximum computational time for each MILP run is set to 1800 seconds.

The comparison of IMLEO over the ten scenarios is shown in Table 6. Also, the optimality gaps of the optimization results, the computation times of the MILP-based methods, and the computation times of the HRL-based methods are presented in Appendix B.

**Table 6 Campaign cost comparison (best-trial)**

| Scenario | Total campaign cost [Mt] | | | | |
| | Stochastic | | | Deterministic | |
| | HRL-PPO | HRL-TD3 | HRL-SAC | MILP (worst) | MILP (semi-worst) |
|---|---|---|---|---|---|
| A | 969.6 | 964.0 | **963.9** | 969.6 | 969.6 |
| B | 1454.5 | 1455.0 | **1432.5** | 1454.5 | 1448.9 |
| C | 1877.9 | **1858.9** | 1861.8 | 1939.4 | 1890.6 |
| D | 2259.3 | 2284.0 | **2249.8** | 2424.1 | 2333.1 |
| E | **1670.8** | 1671.0 | 1756.8 | 1939.3 | 1805.4 |
| F | 2068.5 | **2034.9** | 2068.8 | 2424.1 | 2187.6 |
| G | 1832.9 | **1822.3** | 1879.9 | 1939.3 | 1879.6 |
| H | 2277.0 | **2276.9** | 2303.9 | 2424.1 | 2303.2 |
| I | 3761.4 | 3683.0 | **3619.0** | 3761.4 | 3659.9 |
| J | 3489.2 | **3437.7** | 3443.1 | 3761.4 | 3527.0 |

*Bold numbers indicate the best performance in each scenario.

From Table 6, comparing the results of the MILP-based methods and the HRL-based methods, we can confirm that all solutions provided by the HRL-based methods have lower mean campaign costs than those of the worst and even the semi-worst cases solved by the MILP-based methods for all scenarios. In the proposed scenarios, only two stochastic mission parameters are considered, but it is expected that the HRL-based methods further outperform the MILP-based methods if more stochastic parameters are taken into considerations because the deterministic optimization methods have to take the (semi-)worst situation into account for all stochastic parameters to guarantee the feasibility.

To show the trend of the optimized results, Table 7 provides the detailed mission design; scenario D is chosen as a representative scenario. Multiple sample scenarios of stochastic mission parameter $q$ (ISRU production rate, ISRU decay rate) are picked and their ISRU deployment strategies and spacecraft designs are compared. Note that the ISRU deployment at the first mission and the spacecraft design are the same for all scenarios, as they are determined based on the same initial state. In addition, the deterministic solutions of the worst and semi-worst cases solved by the MILP-based method are also shown in the table, but note that their mission parameters are pre-defined.

**Table 7 Optimized result of ISRU deployment strategy and spacecraft design (Scenario D)**

| Method | mission parameter, $q$ | Spacecraft Design [kg] | | | ISRU deployment [kg] | | | | |
|---|---|---|---|---|---|---|---|---|---|
| | | Payload Capacity | Propellant Capacity | Dry mass | Mis. #1 | Mis. #2 | Mis. #3 | Mis. #4 | Mis. #5 |
| PPO | (2.0, 30) | 3958 | 131479 | 25321 | 5000 | 0 | 0 | 0 | 0 |
| | (5.0, 10) | | | | 5000 | 0 | 0 | 0 | 0 |
| | (8.0, 0) | | | | 5000 | 0 | 0 | 0 | 0 |
| TD3 | (2.0, 30) | 3897 | 129676 | 24960 | 4834.0 | 161.9 | 1623.3 | 1.4 | 0 |
| | (5.0, 10) | | | | 4834.0 | 1.3 | 0 | 0 | 0 |
| | (8.0, 0) | | | | 4834.0 | 0.2 | 0.3 | 0 | 0 |
| SAC | (2.0, 30) | 3790 | 124655 | 24096 | 3545.6 | 1664.3 | 1178.9 | 706.9 | 243.3 |
| | (5.0, 10) | | | | 3545.6 | 1171.5 | 1024.6 | 600.1 | 428.6 |
| | (8.0, 0) | | | | 3545.6 | 895.3 | 657.8 | 471.8 | 431.6 |
| MILP (worst) | (0, 0) | 3723 | 98213 | 20749 | 0 | 0 | 0 | 0 | 0 |
| MILP (semi-worst) | (2.54, 27.3) | 3733 | 102397 | 21278 | 922.5 | 890.5 | 890.5 | 868.4 | 0 |

A few trends can be observed from Table 7: first, we can observe that the HRL-based methods tend to deploy a large amount of ISRU module at the first mission, which is intuitive as the earlier the ISRU module is deployed, the longer time it can produce resources on the lunar surface. On the other hand, the MILP-based method, for the semi-worst case, deploys the ISRU relatively evenly to keep the spacecraft dimension small because the ISRU productivity is precisely known. Second, TD3 and SAC return flexible ISRU deployment strategies in response to different stochastic parameter values $q$. This shows that the infrastructure deployment agent is returning a policy, instead of deterministic action values like MILP. Note that PPO is returning the same values of the ISRU deployment strategies for three stochastic scenarios due to the upper bound. Third, the table also indicates that the spacecraft dimensions designed by the HRL-based methods are larger than those by the MILP-based methods; a similar trend can be confirmed for all ten scenarios. This reflects the robustness of the optimal vehicle design provided by the HRL-based methods; since the deployment of the ISRU modules at the first mission and the spacecraft design are determined without the knowledge of the ISRU parameters, the HRL-based methods deal with this uncertainty by choosing a more conservative mission plan (i.e., larger spacecraft sizing and more ISRU deployment) in this case study.

To examine the reproducibility of the RL algorithms, we choose scenario D as a representative scenario and run 35 trials with the same hyperparameter set for each algorithm so that we can compare the distribution of the optimized results and qualitatively analyze the "trust interval" of the RL algorithms. The box-and-whisker plot for each RL algorithm is shown in Fig. 6. For the other scenarios, similar trends are obtained. Dots in the figure indicate the outliers, which are the data exceeding the 1.5 times of quartile range when extending the whiskers. The deterministic results of the worst and semi-worst cases solved by the MILP-based method are also presented for reference.

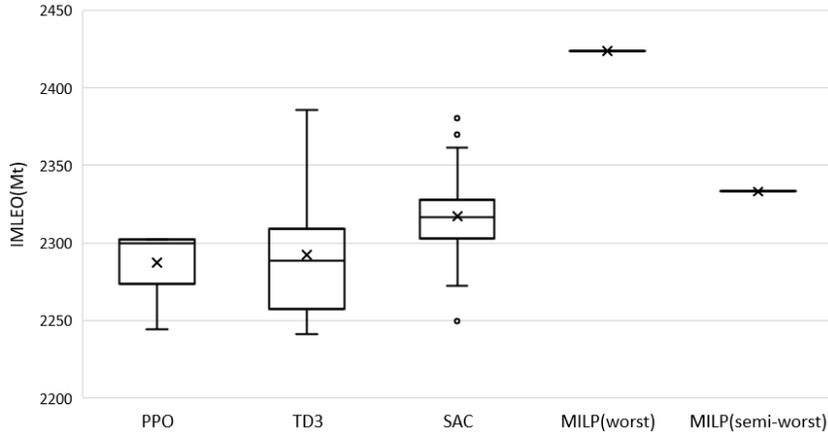

**Fig. 6  Reproducibility of each algorithm (Scenario D)**

Fig. 6 indicates that, while the best trial results (i.e., the smallest IMLEO) have similar values for each RL algorithm, which matches the observation in Table 6, PPO returns relatively reproducible results every run, while the other two algorithms contain larger variances of the total campaign costs. In addition, the averages of the IMLEOs provided by PPO and TD3 are lower than that of SAC. While each RL algorithm has its pros and cons, it is noticeable that all trials for all algorithms had a better performance than the worst result optimized by the MILP-based methods, which validates the general competence of this architecture and the effectiveness of deploying the ISRU modules even with uncertainties under the given condition.

To summarize the findings, we observe that the HRL-based method can successfully capture the flexible infrastructure deployment policy as well as the robust vehicle design. As we include appropriate state variables, the RL agents can learn the mission interdependencies (e.g., future influence of ISRU modules) based on a MDP without running the MILP over a long horizon.

## V. Conclusion

This paper proposes the hierarchical reinforcement learning framework for a large-scale spaceflight campaign design. The particular unique contribution is the developed tri-level hierarchical structure, where three levels of decisions are integrated: vehicle design, infrastructure deployment, and space transportation scheduling. This hierarchical structure enables the RL to be used for the high-level decision and the network-based MILP for the low-level decision, leveraging the unique structure of the space mission design problem for efficient optimization under uncertainty.

The framework is applied to a case study of human lunar space campaign design problems, which include stochastic ISRU production rate and ISRU decay rate. The result is compared with that from the worst-case deterministic scenario, and the effectiveness of the proposed architecture to consider the stochasticity of the parameters is demonstrated. Also, various state-of-the-art RL algorithms for the infrastructure deployment agent are compared and their performances are analyzed.

This paper opens up a new research direction that connects the rapidly growing RL research to the space mission design domain, which was not previously possible due to the enormous action space for the detailed mission decisions. This is achieved by integrating the RL and MILP-based space logistics methods through a hierarchical framework so that we can handle the otherwise intractable complexity of space mission design under uncertainty. Possible future research directions include the methods for more detailed vehicle design, the refinement of the reward definition, or systematic and efficient hyperparameter tuning. Application of Model-based RL based on Partially Observable Markov Decision Process (POMDP) is also considered. It is our hope that this work will be a critical stepping stone for a new and emerging research field on artificial intelligence for space mission design.

## Appendix A. Hyperparameters of RL algorithms

The hyperparameters of each RL algorithm during the training are listed below. These hyperparameters are for Scenario D with Tri-level architecture, and we manually tuned them for the different scenarios and architecture.

*1. PPO*

**Table A1 Hyperparameters for PPO (Scenario D)**

| Hyperparameter | Value |
|---|---|
| Learning rate | 5e-4 (linear) |
| Batch size | 64 |
| Total timestep [episode] | 60 |
| Gamma | 0.95 |
| Lambda (advantage factor) | 0.95 |
| Optimization epoch (gradient steps) | 10 |
| Dual-agent training start [episode] | 15 |
| Network architecture | [64,64] |
| Clip parameter | 0.2 |
| Adam  epsilon | 1e-5 |

*2. TD3*

**Table A2 Hyperparameters for TD3 (Scenario D)**

| Hyperparameter | Value |
|---|---|
| Learning rate | 4e-4 |
| Buffer size | 1,500 |
| Batch size | 64 |
| Tau (soft update of neural network) | 0.005 |
| Total timestep [episode] | 700 |
| Gamma | 0.95 |
| Train frequency / gradient step | 2 |
| Training start (Infra. agent), $n_1$ [episode] | 100 |
| Training start (Vehicle agent), $n_2$ [episode] | 300 |
| Network architecture | [256,256] |

*3. SAC*

**Table A3 Hyperparameters for SAC (Scenario D)**

| Hyperparameter | Value |
|---|---|
| Learning rate | 8e-5 |
| Buffer size | 1,500 |
| Batch size | 64 |
| Tau (soft update of neural network) | 0.002 |
| Total timestep [episode] | 700 |
| Gamma | 0.95 |
| Entropy coefficient | 0.07 |
| Train frequency / gradient step | 2 |
| Training start (Infra. agent), $n_1$ [episode] | 100 |
| Training start (Vehicle agent), $n_2$ [episode] | 300 |
| Network architecture | [256,256] |

# Appendix B. Details of Experiments

*1. Optimality gap and computation time of MILP implementation*

When the MILP optimization is terminated, we tracked the optimality gap between the upper bound and lower bound. The relative optimality gap is calculated as the ratio of the difference between the upper bound and the lower bound to the upper bound (feasible solution) and is shown in Table B2. Also, the computation times for each scenario are shown here as well. Note that the optimality gap is only shown when the optimization is aborted due to the computation time limit.

**Table B1 Computation time and Optimality Gap of the MILP-based method**

| | MILP (worst) | | MILP (semi-worst) | |
|---|---|---|---|---|
| Scenario | Optimality Gap, % | Computation time, s | Optimality Gap, % | Computation time, s |
| A | -- | 6 | -- | 5 |
| B | -- | 134 | -- | 225 |
| C | 23.3 | 1,800 | 22.1 | 1,800 |
| D | 30.6 | 1,800 | 31.2 | 1,800 |
| E | 23.3 | 1,800 | 21.6 | 1,800 |

| | | | |
|---|---|---|---|
| F | 30.6 | 1,800 | 31.9 | 1,800 |
| G | 23.3 | 1,800 | 21.5 | 1,800 |
| H | 30.6 | 1,800 | 34.4 | 1,800 |
| I | 29.9 | 1,800 | 26.6 | 1,800 |
| J | 29.9 | 1,800 | 28.1 | 1,800 |

*2. Computation time of the HRL-based method*

The computation times for each RL algorithm for each scenario are collected as well. Table B3 represents the average computation time spent by the HRL-based method. The computation time only includes the training phase but not the evaluation phase. Additionally, the computation time is determined by the hyperparameters (total timestep in Table A1-A3). Note that purely comparing the computational times between MILP and RL is not fair because the RL performance depends on the hyperparameter values and their tuning is non-trivial; an optimal strategy for hyperparameter tuning/selection is left for future work.

**Table B2 Average computation time of HRL architecture**

| Scenario | PPO, s | TD3, s | SAC, s |
|---|---|---|---|
| A | 663 | 694 | 688 |
| B | 945 | 925 | 913 |
| C | 1,332 | 1,120 | 1,348 |
| D | 1,581 | 1,324 | 1,339 |
| E | 1,378 | 1,220 | 1,201 |
| F | 1,787 | 1,386 | 1,357 |
| G | 1,478 | 1,115 | 1,793 |
| H | 1,736 | 1,315 | 1,358 |
| I | 1,712 | 1,250 | 1,300 |
| J | 1,682 | 1,249 | 1,402 |